\documentclass{article}
\usepackage{spconf,amsmath,graphicx}
\usepackage{color}
\usepackage{xcolor}
\usepackage{CJKutf8}
\usepackage{graphicx}
\usepackage{multicol}

\title{ConvCounsel: A Conversational Dataset for Student Counseling}

\name{Po-Chuan Chen \qquad Mahdin Rohmatillah \qquad You-Teng Lin \qquad Jen-Tzung Chien}
\address{Institute of Electrical and Computer Engineering, National Yang Ming Chiao Tung University, Taiwan}

\begin{document}
%
\maketitle


%
\begin{abstract}
Student mental health is a sensitive issue that necessitates special attention. A primary concern is the student-to-counselor ratio, which surpasses the recommended standard of 250:1 in most universities. This imbalance results in extended waiting periods for in-person consultations, which cause suboptimal treatment. Significant efforts have been directed toward developing mental health dialogue systems utilizing the existing open-source mental health-related datasets. However, currently available datasets either discuss general topics or various strategies that may not be viable for direct application due to numerous ethical constraints inherent in this research domain. To address this issue, this paper introduces a specialized mental health dataset that emphasizes the active listening strategy employed in conversation for counseling, also named as ConvCounsel. This dataset comprises both speech and text data, which can facilitate the development of a reliable pipeline for mental health dialogue systems. To demonstrate the utility of the proposed dataset, this paper also presents the NYCUKA, a spoken mental health dialogue system that is designed by using the ConvCounsel dataset. The results show the merit of using this dataset.
\end{abstract}
\section{Introduction}
\label{sec:intro}
{Student mental health is a sensitive issue as students often face unique challenges. Current mental health services in schools and universities encounter two major issues. First is the high demanding, and second is the imbalanced student-to-counselor ratio. According to the American School Counselor Association (ASCA), the recommended ratio is 250:1 \cite{scr}. Studies have shown that reducing this ratio significantly improves educational outcomes, such as the higher student grade point averages (GPAs) with better discipline \cite{impacts1, impacts2}. However, achieving this ratio requires substantial manpower, and only a few U.S. states meet this requirement \cite{gagnon2016most}.} In addition, {this imbalance negatively impacts both students and counselors. For students, long waiting times (basically exceeding three weeks in some cases in Taiwan) can worsen their mental health conditions. For counselors, excessive workloads increase stress and reduce treatment effectiveness. For instance, a typical session at National Yang-Ming Chiao Tung University (NYCU) may require 12 or more consultations. Moreover, traditional counseling methods may not meet the expectations of new-generation students, who are more accustomed to interacting with digital systems. Researchers suggest that institutions may leverage AI to enhance the counseling process and improve the students' health \cite{ai}.}

{However, the existing datasets for mental health dialogue systems mostly focus on the empathetic responses in open-domain conversations \cite{empathetic, iemocap} or are designed as the emotional support datasets (ESD) covering the strategies such as affirmation, reassurance, suggestions, and emotional reflection \cite{esd1, esd2}. In real-world counseling, the most critical skill is not offering the advice but listening and understanding, avoiding the role of a real counselor. Additionally, students’ psychological issues are unique and cannot be generalized to the other mental health domains. Therefore, this study introduces the ConvCounsel dataset, specifically designed to build a student counseling dialogue system \cite{10011569} centered on the active listening strategy. Active listening is a key technique that encourages clients to express their thoughts, making them feel understood and validated. This dataset includes both speech and text data, consisting of speech data crawled from streaming media and 40 recorded counseling sessions between counselors and students at NYCU, each annotated with precise timestamps and emotion labels by professional counselors.}

{To evaluate the effectiveness of this dataset, we developed the NYCUKA system \cite{nycuka,nycuka2} as an active listener in a mental health spoken dialogue system. This paper further enhanced the user experience by incorporating a visual agent using an animated character, creating the impression that the user is sharing their story with a real person. The proposed system responds to user inputs in Chinese and provides empathetic responses based on counselor-designed prompts and dialogue history, thereby enriching the counseling experience.}

\begin{figure}[th]
    \centering
    \includegraphics[width=1\linewidth]{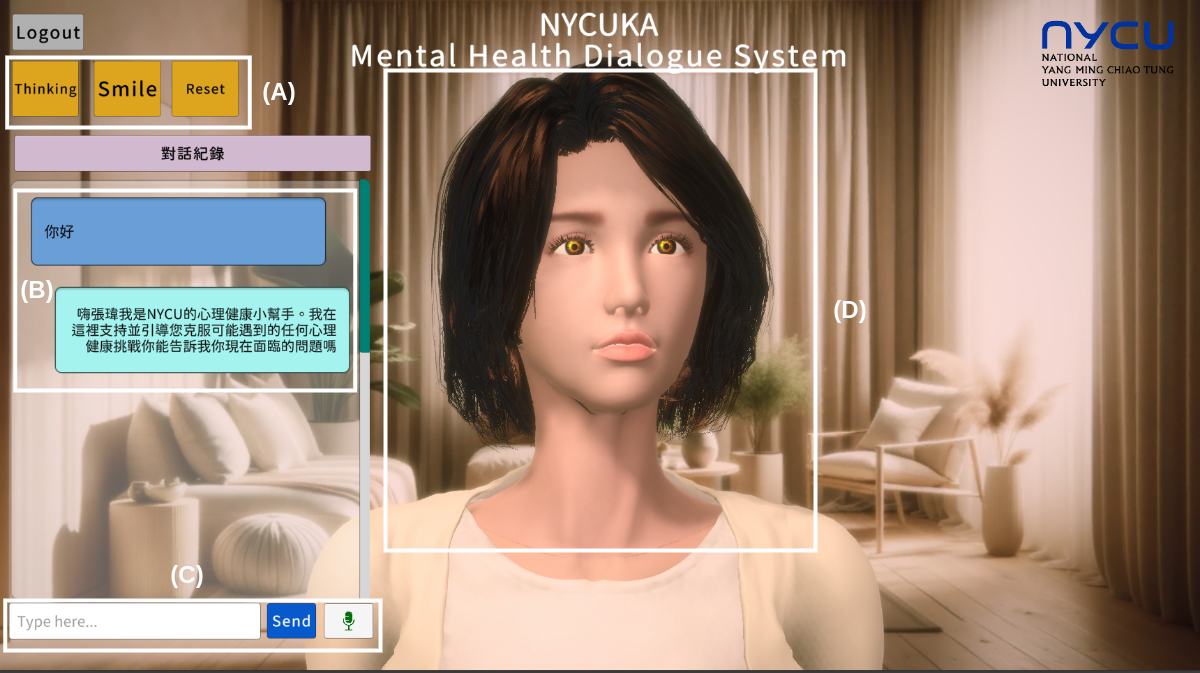}
    \caption{NYCUKA interface consisting of different components which include (A) expression switch for animated character (B) dialogue box (C) user input (either text or voice) (D) animated character}
    \label{fig:animation}
\end{figure}

\begin{figure*}[th]
    \centering
    \includegraphics[width=0.85\linewidth]{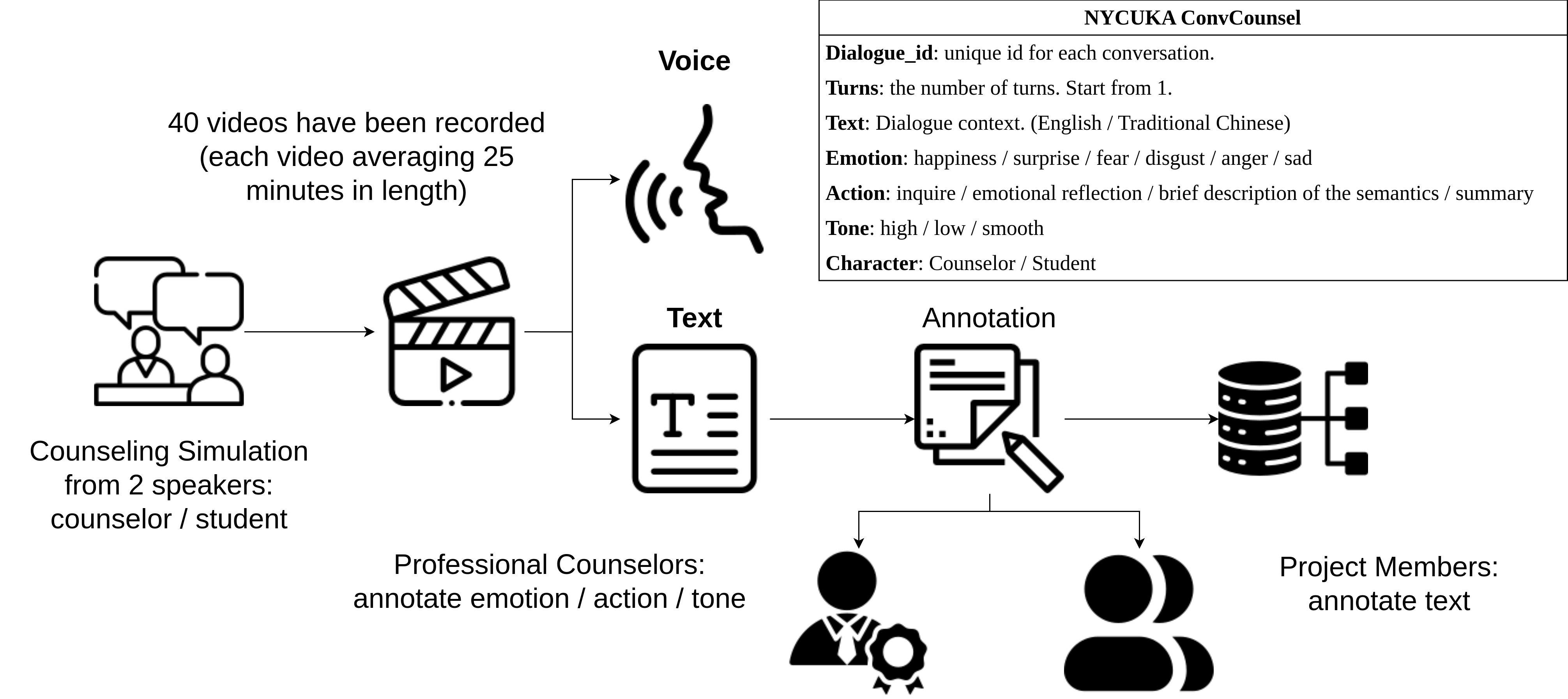}
    \caption{Procedure to conduct data collection for ConvCounsel dataset.}
    \label{fig:collection}
\end{figure*}

\section{Mental Health Dataset}
\label{sec:background}
{Extensive efforts have been made to build mental health-related datasets, with one of the most notable being the EmpatheticDialogues (ED) dataset \cite{empathetic}, which focuses on collecting empathetic responses to general mental health issues. Subsequent datasets \cite{esd1,esd2} expanded on ED by adding labels and annotations for different strategies such as self-disclosure, affirmation, reassurance, and suggestion. However, these datasets have limitations in the context of university counseling. First, general domain datasets like ED did not adequately address the unique issues in the educational sector. Second, learning various strategies in these datasets imposed the challenges for implementing AI-based solutions in the medical field. For example, the dataset in \cite{esd1} included the suggestion strategies at a rate of approximately 16.1\%, which may not be practical since AI cannot provide explicit suggestions like a human counselor. Therefore, this paper proposes a new dataset specifically designed to support mental healthy for university students, emphasizing active listening strategies to effectively elicit student self-disclosure, alleviate stress, and identify appropriate counseling strategies.}

\section{ConvCounsel Dataset}
\label{sec:collection}
The data collection process was conducted in two distinct stages. In the first stage, speech data were obtained through the extraction of videos from a video streaming platform. In the second stage, text data pertinent to the active listening strategy, which is recognized as one of the most crucial counseling strategies, were gathered in collaboration with counselors. The collection process is elaborated as follows.

\subsection{Speech Data}
\label{sec:speech}
{This dataset was constructed by crawling YouTube videos with traditional Chinese subtitles, focusing on speech signals and captions. Data collection was divided into two stages: first, 208 hours of general-domain news data were gathered to cover common vocabulary, followed by 192 hours of domain-specific data from psychology and mental health channels. Speakers from diverse backgrounds were included to ensure robustness to various accents and speaking styles. Both datasets were merged in the collection. The preprocessing techniques, such as data cleaning and downsampling, were applied to ensure consistency and reduce processing load.}


\subsection{Text Data and Annotation}
\label{sec:text}
{The text data were gathered through collaboration with NYCU counselors who conducted realistic simulations of counseling sessions between university students and counselors. These sessions covered various topics based on four strategies including Inquiring, Emotional Reflection, Paraphrasing, and Summarizing. Each session lasted 20 to 30 minutes, discussing issues such as academic challenges, family problems, social relationships, and economic concerns. The dataset includes 40 videos categorized into themes: Self Issues (14 videos), Career Plans (5 videos), Interpersonal Relationships (10 videos), Romantic Relationships (7 videos), Life Adaptation (3 videos), Emotions (9 videos), Family Issues (5 videos), Crisis Management (1 videos), and Academic Issues (5 videos). This diverse collection provides a comprehensive representation of different counseling scenarios. Figure \ref{fig:collection} illustrates the data collection process, and key statistics are shown in Table \ref{tab:nycuka_text}.}


\begin{table}[th] \centering
\begin{tabular}{|c|c|} 
\hline
number of turns   & 4964 \\ 
\hline
video length & 16\,hours\,25\,minutes\,38\,seconds \\ 
\hline
number of themes& 9 \\ 
\hline
\end{tabular}
\caption{\label{tab:nycuka_text} Summary of the statistics of the collected samples.}
\end{table}

An extensive data preprocessing and annotation phase was undertaken after the data collection. This involved meticulously separating each video into multi-turn conversations, systematically identifying and segmenting the dialogues attributed to the counselor and the student. This process resulted in the following annotations in each video.
\begin{enumerate}
    \item \textbf{dialogue\_id:} A unique identifier for each dialogue. All turns that belong to the same dialogue share the same dialogue\_id, enabling the grouping of turns to reconstruct the entire conversation.
    \item \textbf{turns:} An integer representing the order of the turn within the dialogue. Turn one denotes the first turn in the dialogue, Turn two the second, and so forth. This number is annotated to maintain the sequential flow of the conversation.    
    \item \textbf{text:} The actual text of the turn represents the speaker's utterance in that particular turn of a dialogue.
    \item \textbf{emotion:} This label is annotated to categorize the emotional content of the turn. It is a multiclass attribute with 64 possible emotion labels that combine various emotions and intensity levels. The emotions are attributed exclusively to the character (e.g., the student), while the counselor is consistently labeled as ``no emotion''. The six base emotions are happiness, surprise, fear, disgust, anger, and sadness.
    \item \textbf{act:} This label describes the strategy employed by the counselor. It is a multiclass attribute, meaning multiple actions can be presented. This label is annotated solely by the counselor by including four possible actions, namely inquiry, emotional reflection, a brief description of the semantics, and summary, along with their combinations. For the student character, this label is always annotated as ``no action''.
    \item \textbf{character:} This label is annotated to show the speaker's role of the turn, with two possible characters ``student'' and ``counselor''.
\end{enumerate}

In addition, this dataset provides a rich resource for analyzing dialogue structure, emotional expression, and counselor responses in various conversational contexts. This dataset is valuable for applications in natural language processing, conversational AI, and emotion detection, offering the insights into dialogue dynamics and interaction patterns.

\begin{figure}[ht]
    \centering
    \includegraphics[width=0.9\linewidth]{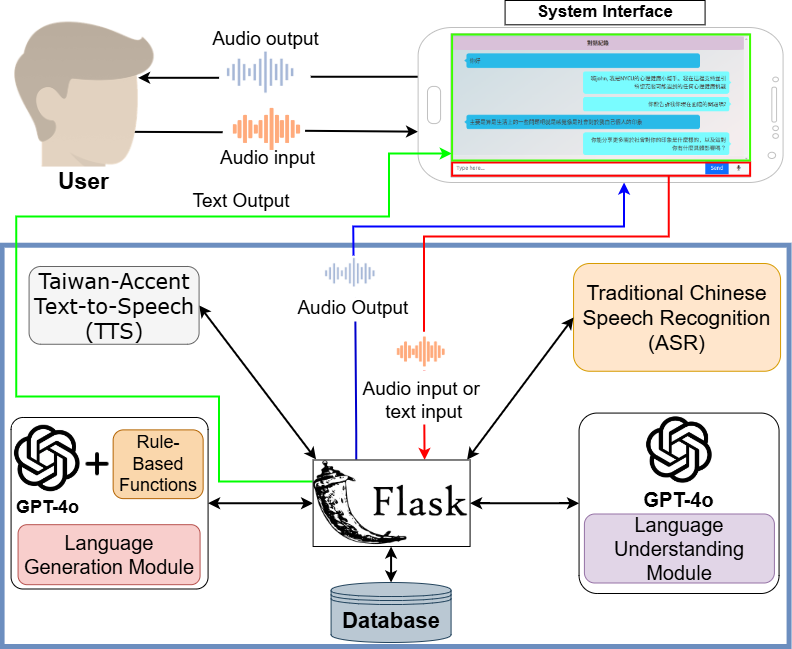}
    \caption{Implementation for NYCUKA dialogue system.}
    \label{fig:nycuka}
\end{figure}

\subsection{Active Listening Strategy}
{The ConvCounsel dataset was collected with the active listening strategy by following the professional counseling techniques including Inquiring, Emotional Reflection, Paraphrasing, and Summarizing. Each technique plays a crucial role in engaging with the student's emotional and psychological state. For example, the Inquiring technique uses the open-ended questions like ``When did you first notice this feeling of anxiety?'' to explore underlying emotional causes. Emotional Reflection mirrors the student's emotions, such as responding with ``It sounds like you're feeling frustrated,'' to foster a supportive environment. Each conversation turn was annotated with the student’s emotions and the counselor's techniques, enabling AI systems to model effective counseling strategies and understand emotional dynamics.}

\subsection{Ethical Considerations}
{The research project has been approved by the National Yang Ming Chiao Tung University Human Research Ethics Committee (IRB number: NYCU112097BE). We have carefully considered privacy and cultural sensitivity with the informed consent. All participants were anonymous, and the consent forms have been signed. Enhanced data security measures are in place to protect privacy. Participants have the right to withdraw their data at any time without any negative consequences. The informed consent process has been approved, and the continuing consent is maintained throughout data collection. To ensure cultural diversity, this dataset contains participants from various backgrounds and was consulted from cross-cultural psychology experts to ensure cultural sensitivity. We are committed to mitigating bias in AI and maintaining transparency regarding the model’s capabilities.}

\section{Experiments}
\label{sec:evaluation}
This dataset was considerably utilized to train the NYCUKA, an active listener dialogue system designed to emulate the counselor responses. The main architecture of NYCUKA is shown in Figure \ref{fig:nycuka}. The primary objective of NYCUKA is to support students with mental issues and help them alleviate their issues before having an in-person counseling session with real counselors. The evaluations were carried out separately between speech and text components.

\subsection{Speech Component}
Firstly, the evaluation was done to evaluate the automatic speech recognition (ASR) model. The model is a Whisper \cite{whisper} miniature version, fine-tuned by using the collected speech data. The connectionist temporal classification (CTC)-based ASR model based on a transformer architecture \cite{asr} was used as a baseline model. The fine-tuned ASR models were evaluated by using a mental health dataset crawled from speech data and their captions from YouTube videos. The training sets in general domain and mental health domain contained 208 hours and 192 hours, respectively, which were used to evaluate the models.

Table \ref{tab:asr_result} reports the Chinese character error rates (CERs) by using different fined-tuned ASR model. Notably, the fine-tuned Whisper model performed much better than the fine-tuned CTC/attention model. The training data in the mental health domain are obviously helpful for ASR performance due to the incorporation of longer training data and closer training domain.

\begin{table}[th] \centering
{\begin{tabular}{|c|c|c|} \hline
training domains & CTC/Attention & Whisper \\ \hline
general    &   13.6 \%           & 8.3 \%                   \\ \hline
general \& mental health         &   6.5 \%             & 3.8 \%        \\ \hline
\end{tabular}}
\caption{\label{tab:asr_result} Results of CERs for ASR by using two fine-tuned models in general and mental health domains.}
\end{table}

\begin{table*} \centering
\small{ \begin{tabular}{|c|c|c|c|} \hline
Student input & Counselor response &GPT-4o with standard prompt & NYCUKA ConvCounsel \\ \hline
  \begin{tabular}[c]{@{}l@{}} It also feels a bit like \\you're afraid of \\the others discovering \\that you’re not \\capable enough, or \\something like that.\end{tabular}  & \begin{tabular}[c]{@{}l@{}}So it seems like this is what \\ you're worried about. It seems \\ like you're saying you're more \\ worried that your abilities \\ aren't sufficient. So, you’ve \\ discovered that this worry is \\ making you are prone to \\ anxiety. When did you first  \\ notice that you were feeling \\ this way?\end{tabular} &  \begin{tabular}[c]{@{}l@{}} The feeling you mentioned is \\ actually quite common. This \\ phenomenon is known in \\ psychology as ``Impostor Syndrome''. \\ $\dots$ But there are strategies to help \\ you cope: 1. recognize and accept \\ this feeling $\dots$ 2. objectively assess \\ your abilities $\dots$ You can \\ always talk to me, and I'll \\ do my best to help you. \end{tabular}   &   \begin{tabular}[c]{@{}l@{}}Hmm, it sounds like you're also \\ worried about others discovering \\ that you’re not capable enough. \\ This worry makes you feel \\ particularly anxious in situations  \\where you need to demonstrate \\ your abilities. $\dots$ Do you \\ feel that your abilities are \\ being questioned, or are you \\ starting to doubt your \\ own abilities? \end{tabular}                   \\ \hline
  \begin{tabular}[c]{@{}l@{}} Yes, even with small \\ responsibilities, I try \\ to share them with \\ the others. \end{tabular}  & \begin{tabular}[c]{@{}l@{}}Is that responsibility to \\maintain the operation of the \\club, or...? \end{tabular} &  \begin{tabular}[c]{@{}l@{}} That's a very wise strategy. \\ Sharing responsibilities not only \\ reduces your stress but also \\ gives other members a greater \\ sense of participation and \\ responsibility. $\dots$ Here are some \\ specific methods to help  you share \\responsibilities more effectively: \\ 1. clear division of labor: \\ define the roles clearly $\dots$ \\ 2. build trust: trust your \\ teammates $\dots$ You can \\ always talk to me.\end{tabular}   &   \begin{tabular}[c]{@{}l@{}}Hmm, it sounds like you're good \\ at finding ways to share \\ responsibilities. This allows \\ you to balance your role \\ as a leader while having \\ more time and energy \\ to do what you enjoy. \\ This arrangement probably \\ makes you feel more \\ balanced and relaxed, right? \end{tabular}                   \\ \hline
\end{tabular}}
\caption{\label{tab:prompt_sample}Examples of using ConvCounsel dataset in NYCUKA by comparing different prompt strategies. This comparison highlights the results of using standard prompts, which primarily offer direct advice, and counselor-designed prompts, which focus on providing emotional support, fostering self-reflection, and responding to the user feelings like a human counselor.}
\end{table*}

\begin{table*}[ht!] \centering
\small{ \begin{tabular}{|c|c|c|c|c|c|} \hline
  & BLEU-1/2 ($\uparrow$) & ROUGE-1/2 ($\uparrow$) & Distinct-1/2 ($\uparrow$) & PPL ($\downarrow$) & BERTScore ($\uparrow$) \\ \hline
standard prompts & 3.91/0.4 & 6.69/1.63 & 30.03/64 & \textbf{12.2196} & 51.72 \\ \hline 
counselor-designed prompts & \textbf{12.98/1.53}  & \textbf{18.4/4.69} & \textbf{69.98/91.8} & 23.1045 & \textbf{58.46} \\ \hline
\end{tabular}}
\caption{\label{tab:text_exp}Evaluation of different metrics by comparing standard prompts and counselor-designed prompts.}
\end{table*}

\begin{figure}[ht]
    \centering
    \includegraphics[width=1\linewidth]{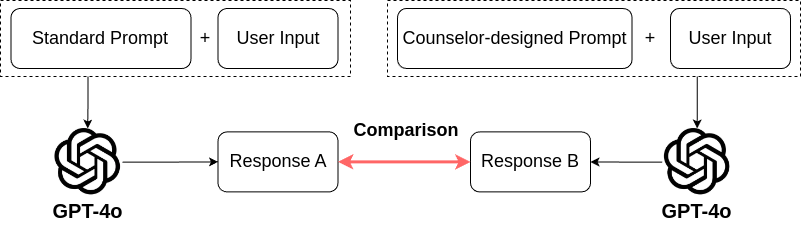}
    \caption{Procedure for comparing the quality of responses generated by GPT-4o using two different prompts.}
    \label{fig:text_exp}
\end{figure}

\subsection{Text Component}
For the text-based evaluation, we conducted a comparison by using a dataset comprising five randomly selected videos from the ConvCounsel. There were two different prompting methods which are the standard prompts and the task-specific prompts designed by NYCU counselors. Both methods utilized the GPT-4o \cite{openai2024gpt4technicalreport} as the backbone large language model (LLM). First, we predefined two distinct prompts as follows
\begin{enumerate}
    \item \textbf{standard prompts}: These prompt were introduced to instruct LLM to act as a counselor to provide the response based on the user input and context.
    \item \textbf{counselor-designed prompts}: These prompts were designed by professional counselors. They were used to guide LLM to classify the user emotions, identify appropriate counseling strategies, and explain the reasoning behind its responses.
\end{enumerate}

{Figure \ref{fig:text_exp} illustrates the effectiveness of the prompt strategies used in the ConvCounsel dataset. Table \ref{tab:prompt_sample} provides some examples from ConvCounsel, highlighting the differences between standard prompts and counselor-designed prompts. The Chinese dialogue has been translated into English. Standard prompts tend to offer direct advice, often introducing psychological concepts or strategies to simulate specific feelings or situations. For instance, in the first example, standard prompts introduce ``Impostor Syndrome'' and provide steps to manage anxiety. Counselor-designed prompts, however, adopt a more empathetic approach, focusing on emotional support, validation, and self-reflection. Under the same scenario, these prompts acknowledge the user's concerns and explore the emotional impact by questioning whether the user feels self-doubt or a lack of confidence. This strategy encourages deepening the conversation, and promoting the emotional processing and expression. Thus, standard prompts are more directive while counselor-designed prompts are more supportive for empathetic dialogue that empowers users.}

{The automated evaluation compares the responses generated with both prompt strategies using GPT-4 under the same settings (temperature = 0, top-p = 1). Responses were assessed against a dataset containing the simulated student contexts with the professional counselor references which were evaluated by
\begin{itemize}
    \item \textbf{Perplexity (PPL)}: This metric measures the model ability to predict the next word in a sequence, indicating the fluency and coherence of the generated text.
    \item \textbf{BLEU-N}: This metric evaluates the quality of the generated text by comparing it with reference text, with BLEU-1 focusing on unigram precision and BLEU-2 on bigram precision.
    \item \textbf{ROUGE-N}: This metric assesses the overlap of unigrams and bigrams between the generated text and the reference texts, providing the insights into the recall and relevance of the generated content.
    \item \textbf{Distinct-N}: This metric measures the diversity of the generated text by calculating the proportion of unique n-grams within the text. A higher Distinct-N indicates greater diversity and less repetition in the output.
    \item \textbf{BERT score}: This metric evaluates the semantic similarity between the generated and reference texts by leveraging contextual embeddings from BERT.
\end{itemize}
Table \ref{tab:text_exp} shows that counselor-designed prompts perform better in BLEU, ROUGE, Distinct, and BERTScore, indicating stronger alignment with reference responses. However, standard prompts achieve lower perplexity, suggesting more fluent, less complex text. This highlights the trade-offs between prompting strategies in text generation tasks.}

{In addition to using automated metrics to evaluate the accuracy and diversity of the system responses, this study also conducted a user satisfaction survey. According to the survey results, the system received an average score of 4.43/5 for response appropriateness, 4.57/5 for emotion recognition accuracy, and 4.36/5 for overall satisfaction. This indicates that the ConvCounsel dataset not only significantly improved the system performance in terms of automated metrics but also received high rating in terms of user experience.}

\section{Conclusions} \label{sec:conclusion}
{This study established the ConvCounsel dataset, explicitly designed for student counseling. The uniqueness of this dataset lies in its emphasis on the active listening strategy, a crucial technique in the counseling process that facilitates user self-disclosure \cite{mahdinnycuka}. The ConvCounsel dataset includes speech and text data covering various issues that students may encounter. Based on this dataset, we developed NYCUKA, a dialogue system that simulates counselor responses. Experimental results show that using this dataset considerably improved the performance of the dialogue system in the domain of student mental health, demonstrating the potential of this dataset in developing reliable mental health dialogue systems.} {Future improvements will focus on incorporating advanced diffusion techniques \cite{yeh2024cross} and modality translation strategies \cite{liu2024modality} to further enhance cross-modality understanding and response generation. One limitation of the current limitation of NYCUKA system is the latency in response generation, particularly in the scenarios requiring multimodal inputs and outputs. Techniques such as fast posterior sampling in conditional diffusion models \cite{Yeh2023} could be extended and employed to address this issue in multimodal response generation.}

\bibliographystyle{IEEEbib}
\bibliography{refs}

\end{document}